\def\year{2018}\relax
\documentclass[letterpaper]{article} 
\usepackage{aaai18}  
\usepackage{times}  
\usepackage{helvet}  
\usepackage{courier}  
\usepackage{url}  
\usepackage{graphicx}  

\usepackage{booktabs}       

\usepackage{subfigure}
\usepackage{lipsum}
\usepackage{amsmath,amssymb,amsfonts,amsthm,mathtools}
\usepackage{enumitem}
\usepackage{multirow, makecell}
\usepackage{multicol}
\usepackage{array}
\usepackage{bm}
\usepackage{color}
\usepackage{amsmath,scalerel}
\usepackage{cases}
\usepackage{capt-of}
\usepackage{diagbox}

\def\reals{\mathbb{R}}
\def\O{\mathcal{O}}
\def\S{\mathcal{S}}

\def\tableFont{\scriptsize}

\DeclareMathOperator*{\minimize}{minimize}

\newcommand{\argmin}{\operatornamewithlimits{argmin}}
\newcommand{\argmax}{\operatornamewithlimits{argmax}}
\newlength{\widebarargwidth}
\newlength{\widebarargheight}
\newlength{\widebarargdepth}

\begin{document}
%
\title{Why Pay More When You Can Pay Less:\\ A Joint Learning Framework for Active Feature Acquisition and Classification}
\author{Hajin Shim \and Sung Ju  Hwang \and Eunho Yang\\
  }

\maketitle
\begin{abstract}
We consider the problem of \emph{active feature acquisition}, where we sequentially select the subset of features in order to achieve the maximum prediction performance in the most cost-effective way. In this work, we formulate this active feature acquisition problem as a reinforcement learning problem, and provide a novel framework for jointly learning both the RL agent and the classifier (environment). We also introduce a more systematic way of encoding subsets of features that can properly handle innate challenge with missing entries in active feature acquisition problems, that uses the orderless LSTM-based set encoding mechanism that readily fits in the joint learning framework. We evaluate our model on a carefully designed synthetic dataset for the active feature acquisition as well as several real datasets such as electric health record (EHR) datasets, on which it outperforms all baselines in terms of prediction performance as well feature acquisition cost.  
\end{abstract}

\section{Introduction}
Deep learning has shown remarkable growth in recent years mainly due to easier access to vast amount of data from the internet, 
and demonstrated significant improvements over classical and standard algorithms on diverse tasks such as visual recognition \cite{6639347,HeZRS16} and machine translation \cite{BahdanauCB14}, to name a few. 

The fundamental assumption for training an accurate deep network is that data is readily available at little or even no cost, such that the model can make predictions \emph{after} observing all available features (in other words, feature acquisition is considered as an independent process against predictions). However, information acquisition is sometimes not only affected by the model (and vice versa) but it also incurs a cost. Consider, for instance, the task of diagnosing a patient for diseases. A human doctor in this case will start the diagnosis by starting with only a few symptoms that the patient initially reported. From there, the doctor will either ask about more symptoms or conduct some medical examinations to narrow down the set of potential diseases the patient might have, until he/she has the enough confidence to make the final diagnosis. Acquiring all features (via all medical tests) in this problem may cause financial burden to patients and more seriously it may increase the risk of not receiving proper treatment at the right time. Furthermore, collecting irrelevant features might add only the noise and make the prediction unstable. 


In this paper, we first formulate the optimization problem with the regularizer based on the feature acquisition cost. As a human doctor diagnoses a patient in the previous healthcare example, we need to decide which unknown features should be discovered in order to be fully confident about our prediction. This process is sequentially repeated until we have collected sufficient but not redundant features. At every examination (or feature acquisition), we pay the pre-defined inspection fee and receive the reward or penalty according to our final prediction.

We then provide the sequential feature acquisition framework with the classifier for predictions and the RL agent for feature acquisitions, in order to systematically solve the proposed optimization problem. Interestingly, it turns out that the classifier in our optimization framework, can be understood as the estimated environment for the RL agent, which is intuitive in the sense that the reward to the RL agent should be based on how confident our classifier is on its final decision. Finally, given new data point with missing entries, our RL agent sequentially chooses features to acquire based on the history. Once the agent decides to end the acquisition phase, the classifier will make a prediction given the acquired features by the RL agent thus far. At the same time, the final rewards are set from the prediction by the classifier, to signal the agent whether the current subset of features is adequate for prediction or not.


We also investigate the effectiveness of information sharing between the RL agent and the final classifier in the joint learning framework. We show that sharing latent features between them gives better performance and corroborate that the feature acquisition and the model should be highly connected.

Our contributions are threefold: 
\begin{itemize}
	\item We formulate the cost-sensitively regularized optimization problem and provide the framework to solve it with the classifier and the RL agent. In our framework, we jointly train the classifier and the RL agent, under which the active feature acquisition problem can be learned in the more systemic and stable way, without requiring given probabilistic model or pre-defined classifier. 
	\item We propose a novel method to encode the subset of features and provide a way to properly handle missing entries shared by the classifier and the agent. In addition, we apply DQN \cite{mnih2013} with the replay memory and delayed update in order to handle real-valued feature space in feature acquisition problems.  
	\item We validate the superiority of the proposed framework on diverse simulated and real datasets. Especially, we empirically show the improvements of recent work \cite{mnih2014recurrent} learning the classifier and the RL agent, by the simple modifications within our framework. 
\end{itemize}


\section{Related Work}

The goal of feature selection is to derive the better generalizations by selecting only the relevant but not redundant features \cite{blum1997selection}. Feature selection has been widely studied for diverse applications to identify the \emph{fixed} subset of features relevant to the target task. It can be done in several different ways including sparsity-inducing regularization~\cite{Tibshirani96}, greedy forward and backward selections~\cite{guyon2003introduction} and reinforcement learning \cite{GaudelS10}.

The active feature acquisition aims to minimize the cost of achieving a desired model accuracy in the training phase. \cite{zheng2002active,Saar-Tsechansky09} for instance estimate the amount of information for the missing features in the training data, and selectively acquire missing entries considering the acquisition cost. On the other hand, in our problem setting, the goal is not to minimize the actual cost spent in the training but to train the models that will make cost-effective predictions in the future (or in test phase).  \cite{Greiner:2002:LCA:635012.635013} investigate the problem of learning optimal active classifier based on a variant of the probably-approximately-correct (PAC) model. \cite{ShengL06} propose a model that acquires a batch of features iteratively until no more positive cost reduction occurs. \cite{kanani2008prediction} identify the data points on which the missing features will be completely acquired based on the expected utility. \cite{bilgic2007voila,yu2009active} approach the cost-effective feature acquisition problems with graphical models. However, all theses models are developed under some stringent assumptions, such that all instances have the same known/unknown features set or have the same acquisition orders.

Starting from the work of \cite{RuckstiessOS11}, reinforcement learning-based algorithms have been developed to more systematically handle the sequential feature selection problem at the \emph{instance-level}. \cite{RuckstiessOS11} formulate a partially observable Markov Decision Process (POMDP) to treat each instance with missing entries as the partial observation of state. However, they assume pre-learned classifier is given and fail to specify when to stop acquiring features. \cite{dulacarnold2011} propose a MDP formulation by directly modeling the state space with acquired features only. They also incorporate the additional special actions corresponding to predictions, which leads to train the classifier (as well as when to stop) implicitly within MDP framework. On the contrary, our framework explicitly train the external classifier that can be potentially any advanced one, more importantly it plays a crucial role in deciding rewards in MDP.  


Another recent line of work \cite{mnih2014recurrent,ba2014multiple,ba2015learning} reduces the computational cost to process the high dimensional image data via the fixed number of recurrent attention, leading to size-independent acquisition costs. \cite{mnih2014recurrent,ba2014multiple} use REINFORCE \cite{williams1992simple} to make the models localize informative part and \cite{ba2015learning} improve these methods with an additional inference network and reweighted wake-sleep algorithm. They train both classifiers and RL agents sharing the input embeddings, however, they consider much easier problem under the assumption that the number of features to be acquired is fixed and known a priori. Those models can be seamlessly extended and interpreted in our framework, leading to better performances as shown in our experiments. 



\section{Sequential Feature Acquisition framework via Reinforcement Learning}\label{Sec:setup}

Consider the standard $K$-class classification problem where we learn a function $f_\theta$ that maps data point $\boldsymbol{x} = (x_1, x_2,\hdots, x_p) \in \reals^p$ with $p$ features to a label $y \in \mathcal{Y} := \{1,2, \hdots, K\}$. The basic assumption here is that the feature vector is fixed-dimensional, and given in full (but possibly with missing entries). We instead consider the same problem under a slightly different constraint. 

For each data point $\boldsymbol{x}^{(i)}$, we actively acquire the features in a sequential order. Specifically, at $t=0$ we start with an empty acquired set $\O_{0} := \emptyset$. 
At every time step $t$, we choose the subset of unselected features, $\S_{t}^{(i)} \subseteq \{1,\hdots,p\}\setminus \O_{t-1}^{(i)}$ and examine the values of missing entries $\S_{t}^{(i)}$ at the cost $\boldsymbol{c}^{(i)}_t := \sum_{j \in \S_{t}^{(i)}} c_j$. Hence, after the examination at time $t$, we have access to the values of $\O_{t}^{(i)} := \S_{t}^{(i)} \cup \O_{t-1}^{(i)}$. We repeatedly acquire features up to time $T^{(i)}$ ($\O_{T^{(i)}}^{(i)}$ is not necessarily equal to all data points $i=1,\hdots,n$) and classify $\boldsymbol{x}^{(i)}$ given only the subset of features $\O_{T^{(i)}}^{(i)}$ observed. Note that the order of feature acquisitions and corresponding costs are different across samples, but we drop the sample index $i$ when it is clear in the context.

\begin{figure}[t]
    \centering
	\includegraphics[width=0.49\textwidth]{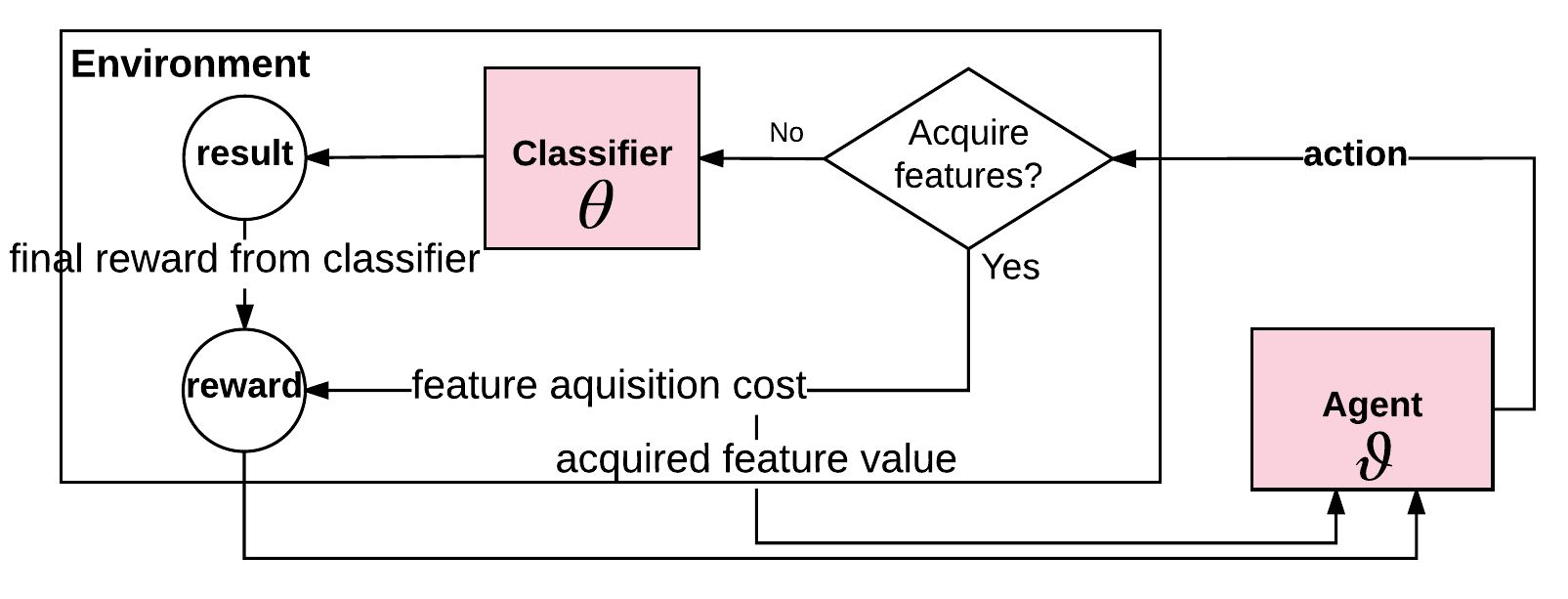}
	\caption{\small Illustration of our reinforcement learning framework for sequential feature acquisition. Each episode corresponds to choosing subset of features to be used for classification. The agent takes an action to choose which information(or feature) to get, and the environment returns the value of it with the feature acquisition cost to the agent, until the agent selects the stop action. At this point, the environment evaluates the quality of chosen features based on the classifier $\theta$ and gives the proper reward to the agent.} 
	\label{fig:concept}
\end{figure}

In order to learn the model that minimizes the classification loss and the acquisition cost simultaneously, we formulate our framework in the following optimization problem:
\begin{align}\label{EqnOurFramework}
	\minimize_{\theta, \vartheta} \ \frac{1}{N}\sum_{i=1}^N \mathcal{L}\Big( f_\theta \big( \boldsymbol{x}^{(i)}, \boldsymbol{z}^{(i)}_\vartheta \big) , y^{(i)} \Big) +  \boldsymbol{c}^\top \boldsymbol{z}^{(i)}_{\vartheta}
\end{align}
where $\mathcal{L}$ is the pre-defined loss function and $\boldsymbol{z}_{\vartheta} \in \{0,1\}^p$ is indicating whether each feature is acquired at the end (or at $T^{(i)}$) when the sequential selection is performed by policy $\vartheta$. Note also that the classifier $f_\theta$ is able to access only available features with $[\boldsymbol{z}_\vartheta]_j=1$. In this framework  (Eq.\eqref{EqnOurFramework}), the optimal parameters of classifier ($\theta$) and selection policy ($\vartheta$) can be obtained by the alternating fashion. Throughout this section, we will show that solving \eqref{EqnOurFramework} with respect to $\vartheta$ can be achieved by the deliberate construction of reinforcement where the reward for the RL agent is based on the given $\theta$ (as shown in Figure \ref{fig:concept}).


\subsection{RL construction for solving  Eq.\eqref{EqnOurFramework} w.r.t. $\vartheta$}
\paragraph{State.} Since the informative features are different across classes, the subset of features our RL agent should select will differ across data points. Without having any prior information on true class, the importance of the missing features can be estimated from the currently available features $\O_t$. To this end, we construct the state $\boldsymbol{s}_t$ as the concatenation of $\boldsymbol{z}_t$ and $\boldsymbol{x}_t$ where the $j$-th entry of $\boldsymbol{x}_t$, $[{\boldsymbol{x}_t}]_j$, is set to zero if $j \notin \O_{t}$ or to the value of the $j$-th feature otherwise. Here, $\boldsymbol{z}_t \in \{0, 1\}^p$ is similarly defined as above, indicating which features are acquired until time $t$: $[\boldsymbol{z}_t]_j = 1$ means $j$-th feature is examined in the past (i.e. $j \in \O_{t}$), and $[\boldsymbol{z}]_j = 0$ means $j$-th feature is undiscovered yet. 


\paragraph{Action.} The RL agent selects which features to examine. The set of all possible actions is simply defined as the power set of $\{1,\hdots, p\}$ (Note that it includes the empty set $\emptyset$, which means to stop acquiring any more features). 
Throughout the paper, we mainly assume that the agent gets one feature at a time only for simplicity, hence the sizes of action space is $p + 1 = |\{ 1, \hdots, p, \emptyset\}|$ under this assumption. Moreover, at time $t$, some actions would be \emph{invalid} if the corresponding features have already been selected previously. $\emptyset$ is a special action that is valid at any time. If an RL agent chooses $\emptyset$, then we stop discovering unknown features and make a prediction based on currently state $\boldsymbol{s}_t$. 


\paragraph{Reward and environment.} We naturally define the reward as the negative acquisition cost. Specifically, in the episode $(\boldsymbol{s}_0 , a_0, r_1, \boldsymbol{s}_1, \hdots, \boldsymbol{s}_T, a_T=\emptyset, r_{T + 1}, \boldsymbol{s}_{T+1})$, $r_{t+1}$ is set as $-c_{a_t}$ for all actions except $\emptyset$. Here, rewards are pre-defined and known even to our RL agent. Note that it is still not trivial since the state transition from $(\boldsymbol{x}_{t-1}, \boldsymbol{z}_{t-1})$ to $(\boldsymbol{x}_{t},\boldsymbol{z}_{t})$ is deterministic, but $(\boldsymbol{x}_{t},\boldsymbol{z}_{t})$ is unknown to the RL agent before actually observing the acquisition in time $t$. 


 Contrary to `feature acquisition' actions, the state transition by $\emptyset$ action is trivial since no further feature values will be revealed: $(\boldsymbol{x}_{t-1},\boldsymbol{z}_{t-1})=(\boldsymbol{x}_{t},\boldsymbol{z}_{t})$. On the other hand, defining a reward is quite challenging. A reward given by the ``environment" for $\emptyset$ should measure how sufficient information has been provided so far for a prediction. Regarding this, suppose that we have an imaginary classifier (or environment) which makes a perfectly accurate prediction as long as the features provided are sufficient. If this classifier fails to correctly classify on some data point $\boldsymbol{s}_t = (\boldsymbol{x}_{t},\boldsymbol{z}_{t})$, a negative reward $r_{\texttt{wrong}}$ should be given. Otherwise, an RL agent gets a reward  $r_{\texttt{correct}}$  ($> r_{\texttt{wrong}}$). However, this notion of ``sufficiency" is now completely hidden to our RL agent since we do not have this perfect classifier in our hands. 
 
 Instead, we use our classier $f_\theta$ as a surrogate of oracle and estimate the amount of sufficiency based on the predictions of $f_\theta(\boldsymbol{s}_T)$. Interestingly, if we set the final reward $r_{T+1}$ as $-\mathcal{L}\big( f_\theta ( \boldsymbol{x}_T, \boldsymbol{z}_T ) , y\big)$, finding the best policy $\pi_\vartheta$ is actually solving \eqref{EqnOurFramework} in terms of $\vartheta$ with classifier $\theta$ fixed, as follows:
\begin{align*}
	& \argmax_{\vartheta} \ \frac{1}{N}\sum_{i=1}^N \sum_{t=1}^{T^{(i)}_{\vartheta}+1} r_t\Big(\boldsymbol{s}^{(i)}_{t-1}, \pi_{\vartheta}(\boldsymbol{s}^{(i)}_{t-1})\Big)\\
	= & \argmax_{\vartheta} \ \frac{1}{N}\sum_{i=1}^N  \bigg[ -\mathcal{L}\Big( f_\theta \big( \boldsymbol{s}^{(i)}_{\vartheta} \big) , y^{(i)} \Big)  -\boldsymbol{c}^\top \boldsymbol{z}_{\vartheta}^{(i)}\bigg] \\
	= & \argmin_{\vartheta} \ \frac{1}{N}\sum_{i=1}^N \bigg[ \mathcal{L}\Big( f_\theta \big( \boldsymbol{s}^{(i)}_{\vartheta} \big) , y^{(i)} \Big) + \boldsymbol{c}^\top \boldsymbol{z}_{\vartheta}^{(i)}\bigg]
\end{align*}
where $\boldsymbol{s}_{\vartheta}$ and $\boldsymbol{z}_{\vartheta}$ are respectively the final state and corresponding $\boldsymbol{z}$ by the policy $\pi_{\vartheta}$.

\paragraph{Policy.}
In order to find an optimal policy, we use Q-learning \cite{Watkins92} for our agent. More specifically, we adopt deep Q-learning \cite{mnih2013} to approximate state-action value function over the continuous state space. Following \cite{mnih2013}, we can make this deep Q-learning to be more stable by using replay memory and delayed update of the target network. 
Note that our sequential feature acquisition framework is not restricted to Q-learning, and any other standard policy learning methods such as policy gradient methods, A3C, TRPO are viable options as well. 	

\section{Joint Learning of RL Agent and Classifier}\label{Sec:Framework}

In our framework, we need to learn the state-action value function $Q$ (parameterized by $\vartheta$) and the classifier $C$ (parameterized by $\theta$; also note that we rename it from $f_{\theta}$ to match with $Q$). Since both components share the input $\boldsymbol{s}$, training them simultaneously can be understood as the multi-task learning. Intuitively, $Q$ and $C$ should share certain degree of information between them, since both aims to optimize the single joint learning objective in \eqref{EqnOurFramework}. However, too much sharing could result in reducing the flexibility of each model, and we should find the appropriate level of information sharing between the two.

\begin{figure}[tb]
	\centering
	\scalebox{0.8}{
	\begin{tabular}{c c}
	\hspace{-0.4in}
	\subfigure[]{\includegraphics[width=0.25\textwidth]{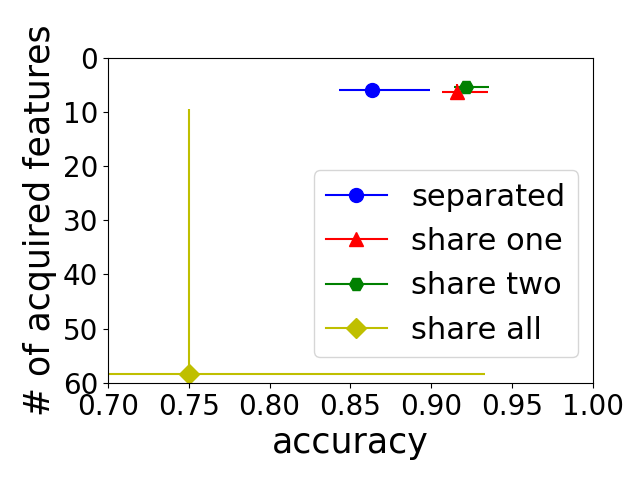}}
	\subfigure[]{\includegraphics[width=0.4\textwidth]{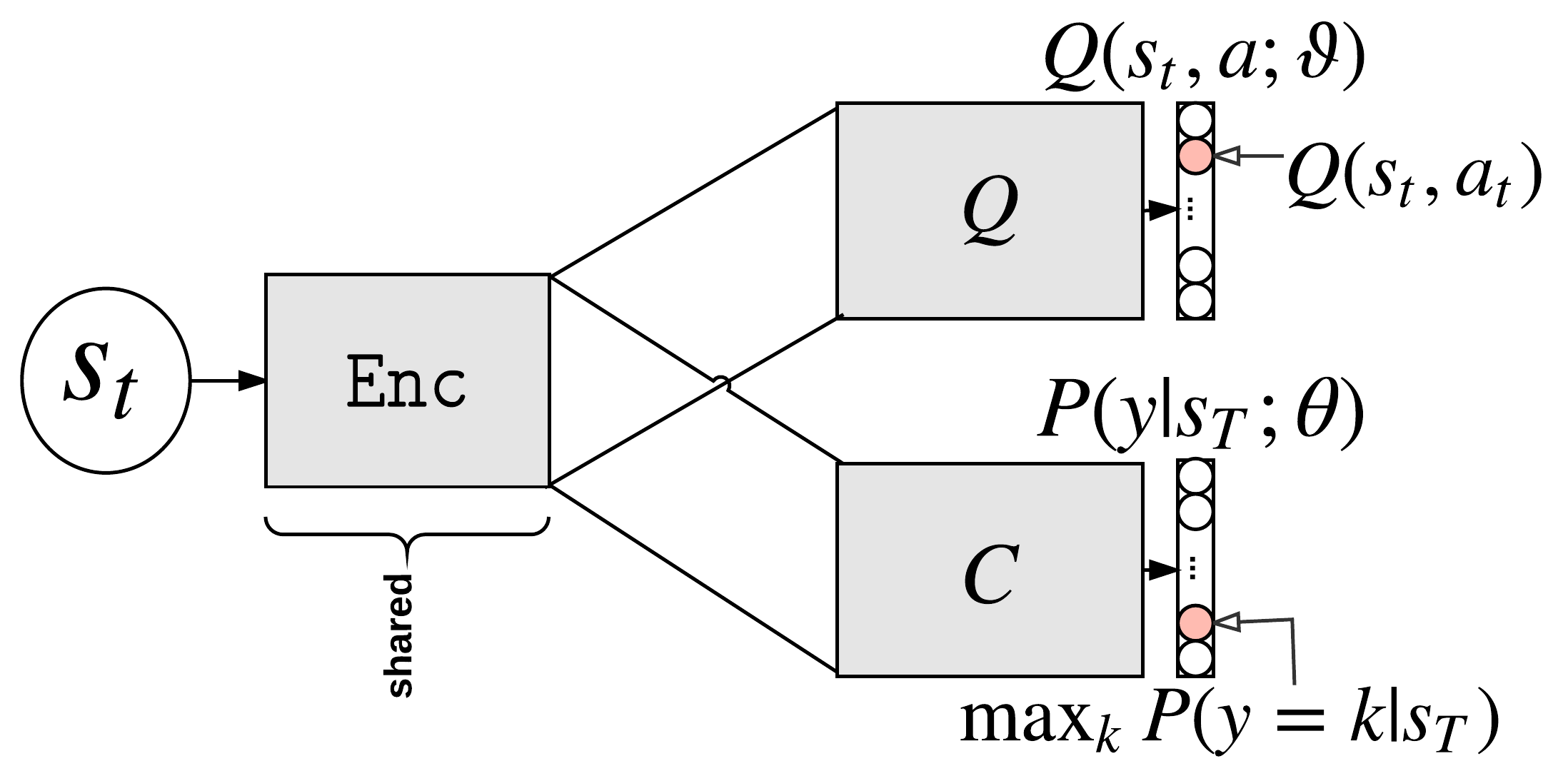}}
	\end{tabular}}
	\caption{\small {\textbf{(a) Effect of sharing:} To see the effect of sharing among $Q$ and $C$, we check the classification accuracy and average number of features collected using our model on the CUBE dataset with 100 features (10 informative and 90 dummy features), with different sharing schemes. Both $Q$ and $C$ are MLP with 3 hidden layers where the sizes are 50-30-50.  We varied the number of shared layers from $0$ (completely separate) to $3$ (completely shared). The points are average accuracy from $100$ runs and error bars represent first and third quartile. (b) Joint learning framework:} Our joint learning framework for RL agent $Q$ and the environment $C$.}
	\label{fig:architecture}
\end{figure}

From our preliminary experiment on the effects of information sharing in Figure~\ref{fig:architecture}(a) (see Section \ref{Sec:sim} for more details about the experimental setup), we found out that the partially sharing model outperforms two other extremes (no sharing or complete sharing) in terms of accuracy and the number of observed features, and also experiences less variance. Motivated by this experiment, we propose the framework of jointly learning the RL agent $Q$ and the classifier $C$ sharing latent features. In our framework, $Q$ and $C$ are semantically separated, but they are allowed to partially share some latent features. For instance, in case where $Q$ and $C$ are multi-layer perceptrons (MLP), they may share the first few layers. These shared layers can also be considered as a shared encoder, $\texttt{Enc}(\boldsymbol{s})$, whose output is fed both into $Q$ and $C$. Hence, our $\texttt{Enc}$-$Q$-$C$ framework to jointly learn $Q$ and $C$ can be formulated in the following way. At every time $t$, the state $\boldsymbol{s}_t := (\boldsymbol{x}_t,\boldsymbol{z}_t)$ is fed to the \emph{shared} encoder:
		$\boldsymbol{h}_t := \texttt{Enc}(\boldsymbol{s}_t)$.
	Then, the encoded representation $\boldsymbol{h}_t$ is given to $Q$ and $C$:
	\begin{align*}
		\boldsymbol{q}_t & := Q(\boldsymbol{h}_t), \quad Q(\boldsymbol{s}_t,a)  := [\boldsymbol{q}_t]_{a} \quad \text{for every action } a , \nonumber\\
		\boldsymbol{p}_t & := C(\boldsymbol{h}_t), \quad	\mathbb{P}(y | \boldsymbol{s}_t) := \texttt{softmax}(\boldsymbol{p}_t) . 
	\end{align*}
The overall architecture of our $\texttt{Enc}$-$Q$-$C$ framework is given in Figure \ref{fig:architecture}(b). While $Q$ and $C$ have their own loss functions, the shared function $\texttt{Enc}$ can be learned both by $Q$ and $C$, or only by one of them, depending on applications. (for example, in learning $Q$, $\boldsymbol{h}_t$ might be considered as a constant, and $\texttt{Enc}$ is trained only through learning of $C$, and vice versa.)   




\subsection{Learning and inference} \label{Sec:Framework_training}
In this subsection we provide the details on how we can actually train $Q$ and $C$ jointly in an end-to-end manner, in our shared learning framework. 
We basically follow the learning procedure described in \cite{mnih2013}, and adopt the two key mechanisms of DQN learning: experience replay memory and delayed update of target Q-network $Q'$ to prevent perturbation. Specifically, in a training phase, 
the agent generates the episode $(\boldsymbol{s}_0, a_0, r_1, \boldsymbol{s}_1, a_1, \hdots, \boldsymbol{s}_{T+1})$ for each data point according to its policy determined by current Q-values. 
For each state, the Q-value of invalid actions are set to $-\infty$. All experience history, $(\boldsymbol{s}_t, a_t, r_{t+1}, \boldsymbol{s}_{t+1})$ for $t=0,\hdots,T$, is saved into the replay memory (if it exceeds the memory capacity, the most recent experiences are retained), so that we can revisit the previous experiences during the training, which makes deep Q-learning more stable by alleviating the dependency of samples.

The immediate reward $r_t$ comes from the environment. For `feature acquisition' actions, the agent gets the pre-defined feature acquisition cost $r_{t} =-c_{a_{t-1}}$. On the other hand, when $a_{t} = \emptyset$, the reward is calculated based on the prediction result from $C$. However, the prediction result from ``premature" $C$ at this point might be noisy, so that the reward calculation is deferred until this history is sampled from replay memory. Therefore in the episode generation phase, this experience tuple is saved in the form: $(\boldsymbol{s}_t, a_t = \emptyset, r_{t+1}=\texttt{undefined}, \boldsymbol{s}_{t+1} = \boldsymbol{s}_t)$. The reward for this experience will be computed (with smarter $C$) when it is sampled in mini-batch and actually used for training. 

After generating episodes for each sample, the mini-batch $\{(\boldsymbol{s}_t, a_t, r_{t+1}, \boldsymbol{s}_{t+1})\}$ is drawn from the replay memory. For experience tuples whose action $a_t=\emptyset$, the rewards are estimated with current $C$ at this point. 

Given the mini-batch for training, all parameters of $Q$ are learned by the gradient decent method to minimize the squared error $(Q(\boldsymbol{s}_t, a_t) - \texttt{target})^2$ where $\texttt{target}$ is $r_t + \max_{a}{Q'(\boldsymbol{s}_{t+1}, a)}$, with the delayed updated $Q'$ for stability. It is worth to note that the discount factor is 1 here since we care about the overall cost for feature acquisitions without discounting.  


While $Q$ is trained, $C$ is also jointly trained. Since $C$ is supposed to perform a classification task with missing values, it would be helpful to train it with incomplete dataset. Toward this, we simulate incomplete data also from the mini-batch of the replay memory. With this mini-batch, $C$ is trained by the gradient descent method to minimize the cross entropy loss: $-\log C_{y_{\texttt{true}}}(\boldsymbol{s}_t)$ where $C_{y_{\texttt{true}}}(\boldsymbol{s}_t)$ is the output (or probability after $\texttt{softmax}$ layer) corresponding to the true label. 
$Q$ and $C$ are alternatively updated until the stopping criteria are satisfied.

\paragraph{Inference.} Once $Q$ and $C$ are trained, we can trivially perform the active feature acquisition for new data points. The start state might be the set of partially known features or a completely empty set. Our RL agent determines which features should be acquired by greedily selecting the action with the maximum Q-value until $\emptyset$ action is chosen. When $\emptyset$ is selected, $C$ makes a prediction based on the features acquired so far.

\subsection{Encoding Data with Missing Features}

As an example of the feature encoding $\texttt{Enc}$ in our joint learning framework, we now describe the set encoding method recently proposed in \cite{Vinyals16}. The \emph{set} encoding is well suited to $\texttt{Enc}$ since it naturally distinguishes between two ambiguous cases: i) $j$-th entry is missing and ii) $j$-th entry is discovered but its value is zero.

The set encoding in \cite{Vinyals16} is composed of two main parts: i) a neural network called $\emph{reading block}$, which maps each element $x_i$ of the input to the real vector $m_i$ and ii) an LSTM called $\emph{process block}$ that processes $m_i$ and repeatedly apply the attention mechanism to produce the final set embedding.  

We adapt this set encoding method to represent each state $\boldsymbol{s}_t$ in the following way: we individually treat the pair of feature index and its value observed, $(j: x_j)$, as the element in the set. Since the actual value of feature index does not convey any information, we first represent each observed feature as $u_j = (x_j,\mathcal{I}(j))$ where $\mathcal{I}(j) = (0, ..., 0, 1, 0, ..., 0)$ is the one-hot vector with 1 for $j$-th coordinate and 0 elsewhere in order to incorporate the coordinate information. Then, via the set encoding mechanism (through the reading block to make $\{m_j\}_{j \in \O_t}$, followed by the process block) introduced above, we produce the set embedding with the observed features.

\section{Experiments}\label{Sec:sim}

To validate the versatility of our feature acquisition model, we perform the extensive experiments both on simulated and real datasets. 

\paragraph{Experimental setup}

Throughout all experiments, we use the Adam optimizer with a fixed learning rate of 0.001 to train our model and train the models until training steps reach to the fixed number of epochs. 

\subsection{Simulated dataset: CUBE-$\sigma$}
To check if the agent can identify few important features that are relevant to the given classification task, we experiment on a synthetic dataset, CUBE-$\sigma$ (See the detailed description of the dataset in \cite{Ruckstiess13}). 
\begin{figure}[tb]
\centering
\includegraphics[width=\columnwidth]{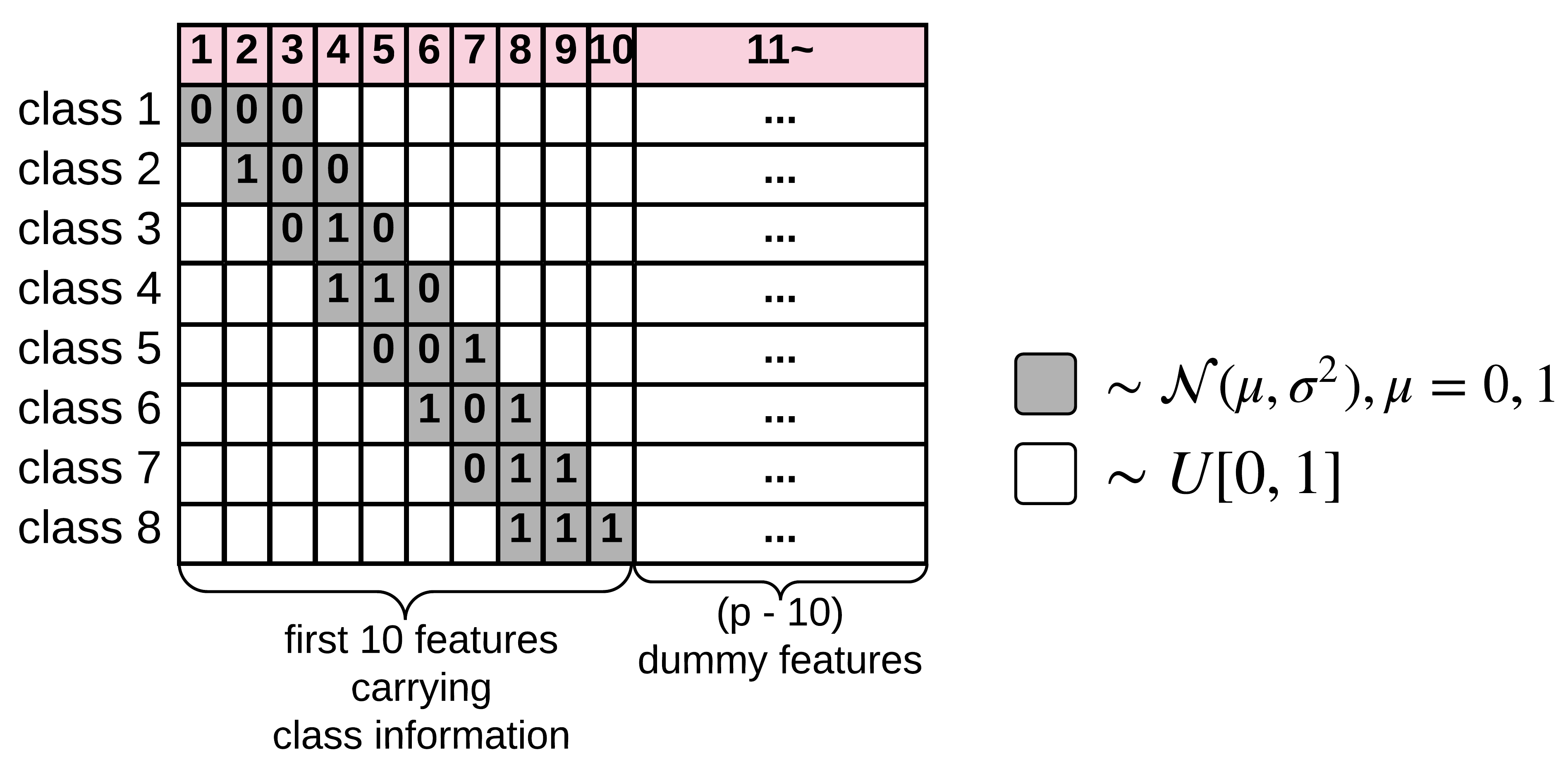}
	\caption{\small The CUBE dataset consists of p-dimensional real valued vectors in 8 classes. The first 10 features only carry class information with three normally distributed entries in the different locations(dimmed in the figure with written mean values) according to the classes. The rest of them are just uniformly random.}
	\label{fig:cube}
\end{figure}

In this dataset, the first 10 features are informative and the other features are dummy features randomly generated from uniform distribution (See Figure~\ref{fig:cube}). The dataset has total of 8 classes, each of which has the class-specific features appearing at different positions (Shaded elements in Figure \ref{fig:cube}). Those informative features are independently generated from normal distribution with the mean $\mu$ (which is 0 or 1 depending on classes) and predefined $\sigma$ in each coordinate. 


We train our model on CUBE dataset with total of 20 features, setting the Gaussian noise level $\sigma=0.1^2$ for the informative features. We train $Q$ , $C$ and $\texttt{Enc}$ for 4 epochs on $10,000$ training instances. We assume the uniform feature acquisition cost for 20 features set as $-0.05$ and the final reward is set as the negative loss $-\mathcal{L}$ computed from $C$. $Q$ has two hidden layers of 100-50 units, $C$ has two hidden layers of 20-20 units, and \texttt{Enc} has 40-30 units with the set embedding vector dimensionality set to 20. For exploration of our RL agent, $\epsilon$ linearly decreases from 1 to 0.1 for the first two epochs. 

The overall results show that the agent can successfully learn which features are informative and hence it only selects the features from the ten informative ones. Further, it consumed only 4.87 features on average while obtaining 95.8$\%$ accuracy (even without fine-tuning), which is comparable to the accuracy of the MLP classifier trained with all 20 features ($96.98\%$).

\begin{table}[t]
	\caption{\small Example episodes. We provide these examples to show how our trained agent acts on the cube dataset. The agent successfully selects the cost-effective subset for both two example cases.}
	\label{episode}
	\small
	\setlength{\tabcolsep}{4pt}
	\centering
	\tableFont
	\begin{tabular}{c||c c c c||c c} 
	\hline
	&\multicolumn{4}{c||}{First episode} &\multicolumn{2}{c}{Second Episode}\\
	\hline
	&Init state &Step 1 & Step 2 & Step 3 & Init state & Step 1\\
	\midrule
	Selected feature&& 7th& 3rd & 5th &  & 7th\\
	Feature value && 0.855 & 0.667 & 0.6796 &  & 1.146\\
	$\mathbb{P}(y=c|s_t)$&0.123&0.169&0.373&0.995 & 0.121 & 0.985\\
	\hline
	%
	%
	\end{tabular}
\end{table}

\paragraph{Analysis of Episodes.} To show our model can learn sequential feature acquisition in an effective and efficient way, here we take a closer look at how our trained model works by examining two specific test episodes summarized in Table \ref{episode}. 

The first episode, where the true class is 8, consists of 3 feature acquisition actions.  Following the greedy policy (choosing the action that has the maximum Q-value), the agent collects 7th, 3rd and 5th features in an order and make a prediction. The value of the 7th feature is 0.855, which informs the model that the class is unlikely to be 6 or 7 whose $\mu=0$ (see Figure \ref{fig:cube}). Next, by acquiring the third feature (which is 0.667), that is highly likely to be generated from the uniform distribution, class 1,2 and 3 are excluded as they all use the third feature. In terms of the certainty of classifier, the probability of true class 8 increases from 0.16 to 0.37 after step 2. The next greedy action is to select the 5th feature (0.6796), which allows the agent to further eliminate class 4 and 5. After these 3 steps, the agent selects the action $\emptyset$, and the classifier gives the answer as `class 8' with probability 0.99, which means that it is almost certain about its this prediction. 

The second episode in Table \ref{episode} is the more extreme case. The agent acquires the 7th feature that is greater than 1. In this case, the agent can be sure that that value is from the normal distribution since it is beyond 1. The only possible answer here is `class 5' where 7th feature generated from the normal with $\mu=1$. Our agent successfully catches this case and predicts the class only after a single feature collection.

\begin{figure}[t]
    \centering
    \begin{tabular}{c c}
    	\hspace{-0.1in}
	\includegraphics[height=3cm]{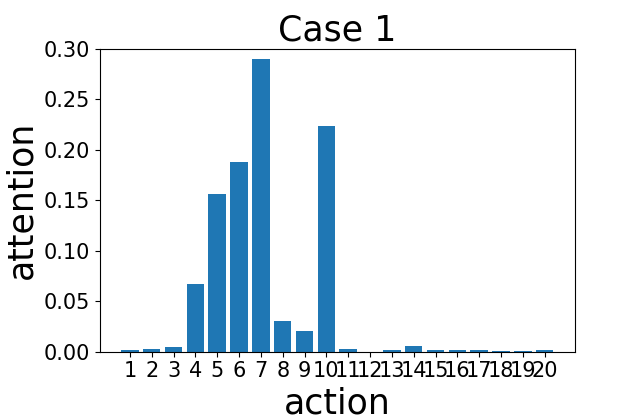}&
	\hspace{-0.2in}
	\includegraphics[height=3cm]{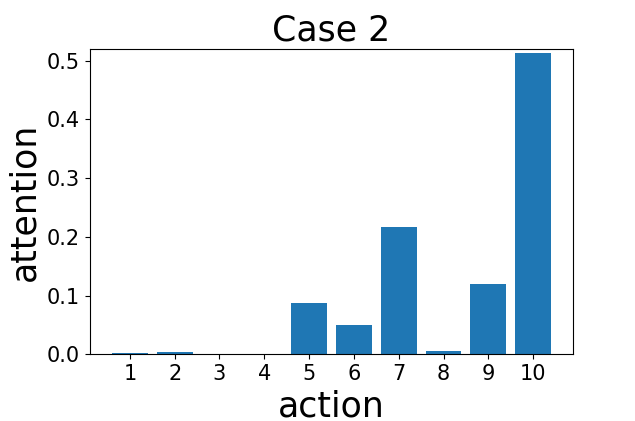}\\
	(a) 1st case & (b) 2nd case\\
	\end{tabular}
	\caption{\small (a-b) Attention on features for two cases. $x$ and $y$ axes represent the feature index and the attention value respectively. Only informative features for classification are being focused on for both cases.}
	\label{fig:attention}
\end{figure}

\paragraph{Analysis of Attention.} The encoder $\texttt{Enc}$ uses the attention mechanism in the process block. In order to examine how it works, we select two typical examples among the cases where the agent observe more than 10 features including dummy features. In the first case (Figure~\ref{fig:attention}(a)), the agent observe all 20 features. Note that this sometimes can happen in our data setting since some data point indeed is quite difficult to be distinguished due to the overlap between uniform and Gaussian distributions (for instance, suppose we have some values close to $(0, 1, 1, 1)$ for 7th-10th features. Then, it is indistinguishable for class $7$ or $8$.). In this case, the agent collects more features, even if it makes no actual difference, since action $\emptyset$ will give low reward in high probability right away while Q-value approximator overestimates the value of the feature acquisition actions. 

As shown in Figure~\ref{fig:attention}(a), the attention mechanism focuses only on the task-related features (3rd to 10th). In the second case (Figure~\ref{fig:attention}(b)), the agent obtains the first 10 features for class 7 data. To differentiate between the class 6, 7, and 8, we need features from 6th to 10th except the 8th feature that are sampled from $\mathcal{N}(1, 0.1^2)$ for all three classes (see Figure~\ref{fig:cube}). We observe that our model allocates attention to only those four important features.
  
\begin{figure}[tb]
\centering
\includegraphics[width=0.9\columnwidth]{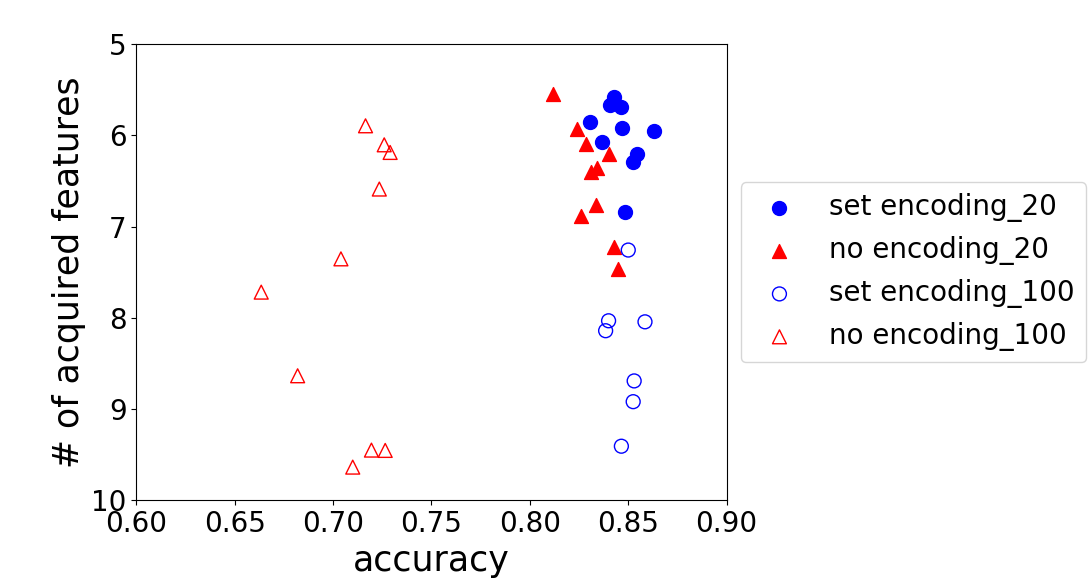}
	\caption{\small The change of accuracy and the number of acquired features on CUBE-0.3 dataset as the number of dummy features increases from 10 to 90. Each point represents the result of a single run. We use the validation set to select the best setting, and repeat the training ten times to obtain the test result. Separate MLP $Q$ and $C$ model shows decreased accuracy and collect larger number of features as the number of dummy features increases, while the accuracy of the set encoding model stays at the same level even with more features.}
	\label{fig:cube_cmp}
\end{figure}
\begin{table}[tb]
	\caption{\small We fix RAM \cite{mnih2014recurrent} to solve \eqref{EqnOurFramework}. We compare modified-RAM against the set encoding as well as the original RAM with the predefined $n=6,8,10$ acquisitions, on CUBE-0.3 dataset with 20 features. We use the cross-validation to choose the models. Results are averaged over 10 trials. Modified-RAM achieves the higher accuracy than the original model with $n=8$ while getting less features.}
	\label{ram}
	\small
	\setlength{\tabcolsep}{4pt}
	\centering
	\tableFont
	\begin{tabular}{c||c|c}
	\toprule
	&accuracy&\# of features\\\midrule
	RAM$(n=6)$& 78.6$\pm$0.007& 6\\
	RAM$(n=8)$& 84.9$\pm$0.001&8\\
	RAM$(n=10)$&86.7$\pm$0.002&10\\\midrule
	modified-RAM&85.4$\pm$0.002&6.77$\pm$0.120\\
	set encoding&85.8$\pm$0.002&7.03$\pm$0.269\\\bottomrule
	\end{tabular}
\end{table}

We also evaluate the use of set encoding in our framework. Given the same classifier $C$ and the RL agent $Q$ fixed in our framework to solve \eqref{EqnOurFramework}, we compare the performances the set encoding against the use of naive encoding: $(\boldsymbol{x}, \boldsymbol{z})$. We again use the CUBE data with 0.3 standard deviation and 20 or 100 features (with more dummies). The scatter plot in Figure~\ref{fig:cube_cmp} shows the accuracy and the number of features acquired for both encodings. The model with the set encoding outperforms both in terms of accuracy and the number of acquired features, showing the resilience against the increase of dummy features (the accuracy of naive encoding decreases with higher variance as the number of dummy features increases from 10 to 90).

\begin{table*}[t] 
\small
	\caption{Results of Physionet 2012 data. We report the test AUC (top) and the average number of acquired features (bottom) by cross validation. For models in our framework, we take the results only among the number of acquired features are less than 10. }
	\label{physio_result}
	\centering
	\tableFont
	\begin{tabular}{l ||c|c|c|c|c||c|c}
		\toprule
	 \multirow{2}{*}{\diagbox[width=10em]{task}{model}}
	 &\multicolumn{5}{c||}{Baselines}&\multicolumn{2}{c}{Models in our framework} \\ \cline{2-8}
	 & MLP & RAM $n=5$ & RAM $n=10$ & RAM $n=20$& no encoding  & set encoding & modified-RAM \\ \midrule  
	\multirow{2}{*}{Mortality}
	&0.816&0.68& 0.74&0.80&0.74&0.81&0.81\\
	& 27.17 & 5 & 10 & 20  & 1.01& 5.30 & 2.07\\ \hline
	\multirow{2}{*}{Length of stay $<$ 3}
	&0.735& 0.69& 0.77 & 0.78& 0.70& 0.74  & 0.70\\
	&27.17& 5 & 10 & 20 & 3.23 &3.9  & 1.14\\ \hline
	\multirow{2}{*}{Cardiac condition}
	&0.919&0.81&0.86&0.91&0.82& 0.87&0.90\\
	&27.17&5&10& 20&6.02& 5.78&2.32 \\ \hline
	\multirow{2}{*}{Recovery from surgery}
	&0.725& 0.61&0.68&0.75&0.63&0.67&0.68\\
	&27.17& 5&10&20&1.01&5.07&3.28 \\ \bottomrule
	\end{tabular}
\end{table*}
\begin{table}[t]
	\caption{Statistics of Physionet 2012 data (\# false: \# true).}
	\label{physio_stat}
	\centering
	\tableFont
	\begin{tabular}{l ||c|c|c}
		\toprule
	  \diagbox[width=10em]{task}{data}& train & validation & test \\\midrule
	Mortality&2586~:~414&443~:~~57&417~:~~83\\
	Length of stay $<$ 3 &2917~:~~83&485~:~~15&487~:~~13\\
	Cardiac condition&2333~:~667&401~:~~99&392~:~108\\
	Recovery from surgery&2209~:~791&363~:~137&360~:~140\\\bottomrule
	\end{tabular}
\end{table}

\paragraph{Fixing RAM in our framework} Recurrent attention model (RAM) and its variations DRAM, WS\_RAM \cite{mnih2014recurrent,ba2014multiple,ba2015learning} reduce the computational cost while obtaining high performance by taking only the informative parts of an image sequentially as a form of glimpse. Especially, RAM and DRAM use REINFORCE algorithm to make the models localize informative parts. However, they take glimpses for predefined $n$ steps for all instances. It may lead redundant consumption for some cases. We fix RAM in the form of our framework, by augmenting the action space with $\emptyset$ action, modifying the rewards based on the classifier, and compare \emph{modified-RAM} with the original on CUBE-0.3 with 20 features. For the feature acquisition problem, the glimpse can be given as $u_j = (x_j,\mathcal{I}(j))$. We can observe in Table~\ref{ram} that the modified-RAM in our framework solving \eqref{EqnOurFramework} achieves higher accuracy than the original models obtaining more features.

\subsection{Case study: Medical Diagnosis}
As the main motivation of our work was to perform cost-effective prediction in the medical diagnosis process, we further examine how our model operates on the EHR dataset from Physionet challenge 2012 \cite{PhysioNet}. 
Physionet challenge 2012 dataset is a time-series data of 48 hours' activities for 8000 patients in the Intensive Care Unit (ICU). It consists of 33 features including \emph{Albumin}, \emph{heart-rate}, \emph{glucose}. We considered the 4 different binary prediction tasks, namely in-hospital mortality, whether length-of-stay was less than 3 days, whether the patient had a cardiac condition, and whether the patient was recovering from surgery, following the experimental setup of \cite{che2016recurrent}. We only use the features in the last timestep, as our model is implemented as a feedforward network. Using only the instances whose the mortality label are publicly available, we randomly split the set into training set, validation set and test set by 3000:500:500 ratio. The data is imbalanced (see Table~\ref{physio_stat}), hence we use weighted cross entropy as the objective function for the classifier $C$. 

For comparison, we train MLP (with full features), RAM, DWSC \cite{dulacarnold2011}, Q-only model (same with DWSC but trained by DQN) and the separate $Q$, $C$ model without $\texttt{ENC}$ as baselines and the models in our framework such as modified-RAM and the MLPs with the set encoder. DWSC and Q-only are the models whose action space includes not only the feature acquisition but also the classification actions. We omit the results of these two models in Table~\ref{physio_result} because they failed to handle the imbalanced data and just learn to predict as the majority. To see whether the informative features are selected, we filter out the results that use more than 10 features on average. Table~\ref{physio_result} shows the AUC and the number of acquired features on the test data with hyperparameters tuned through cross-validation. Set encoding and modified-RAM models achieve AUC close to that of MLP obtained on all features. This shows that our models are able to select informative, task-related features well. 

Our models recovers several characteristics of in-hospital mortality predictions. Among 33 features, the first feature selected is \emph{Glasgow coma scale (GCS)}, which represents the level of consciousness. GCS is a decisive feature as having very low GCS means that the patient is almost unconsciousness. Thus in such a case, the agent stop examining and the classifier predicts that mortality is true. The other major features selected are \emph{blood urea nitrogen}, \emph{serum creatine}, \emph{gender} and \emph{age}.  

\section{Conclusion}
A cost-aware sequential feature selection can be used in situations where the features are not provided in full and each collection of features incurs variable cost, such as with medical data. To solve this problem, we formulated it into an optimization problem of simultaneously minimizing the prediction loss and the feature acquisition costs, and derived a joint learning framework for the classifier and the RL agent. Our model sequentially collects a feature considering its usefulness for prediction and the collection cost, and performs prediction using only a fraction of the original set of features. Specifically, we design our model into a multi-task deep network that is jointly trained for both learning the optimal policy and the classifier, that shares the lower network layers for encoding the set of features that have been collected. We validate our model on both synthetic and real medical data for classification, against relevant baselines, on which it significantly outperforms the baselines using only a number of features, often obtaining even better performance than the model that has access to the full set of features. As future work, we plan to further apply our sequential feature selection model for more cost-critical prediction problems in medical domains.

\fontsize{9.5pt}{10.5pt}
\bibliography{feature_acquisition}
\bibliographystyle{aaai}
\end{document}